\documentclass{article}

\usepackage{arxiv}

\usepackage[utf8]{inputenc} 
\usepackage[T1]{fontenc}    
\usepackage{hyperref}       
\usepackage{url}            
\usepackage{booktabs}       
\usepackage{amsfonts}       
\usepackage{nicefrac}       
\usepackage{microtype}      
\usepackage{lipsum}		
\usepackage{graphicx}
\usepackage{natbib}
\usepackage{doi}

\usepackage{amsmath}

\title{Weather-Dependent Variations in Driver Gaze Behavior: A Case Study in Rainy Conditions}

\author{ 
    Ghazal Farhani \\
	Connected \& Automated Vehicles\\
	National Research Council Canada\\
	London, Ontario, Canada \\
	\texttt{ghazal.farhani@nrc-cnrc.gc.ca} \\
	\And
	Taufiq Rahman \\
	Connected \& Automated Vehicles\\
	National Research Council Canada\\
	London, Ontario, Canada \\
	\texttt{Taufiq.Rahman@nrc-cnrc.gc.ca} \\
    \And
    Dominique Charlebois \\
	Transport Canada\\
	Ottawa, Canada \\
	\texttt{dominique.charlebois@tc.gc.ca} \\
}

\renewcommand{\shorttitle}{\textit{arXiv} Template}

\hypersetup{
pdftitle={A template for the arxiv style},
pdfsubject={q-bio.NC, q-bio.QM},
pdfauthor={Ghazal Farhani, Taufiq Rahman},
pdfkeywords={First keyword, Second keyword, More},
}

\begin{document}
\maketitle

\begin{abstract}
	Rainy weather significantly increases the risk of road accidents due to reduced visibility and vehicle traction. Understanding how experienced drivers adapt their visual perception through gaze behavior under such conditions is critical for designing robust driver monitoring systems (DMS) and for informing advanced driver assistance systems (ADAS). This case study investigates the eye gaze behavior of a driver operating the same highway route under both clear and rainy conditions. To this end, gaze behavior was analyzed by a two-step clustering approach: first, clustering gaze points within 10-second intervals, and then aggregating cluster centroids into meta-clusters. This, along with Markov transition matrices and metrics such as fixation duration, gaze elevation, and azimuth distributions, reveals meaningful behavioral shifts. While the overall gaze behavior focused on the road with occasional mirror checks remains consistent, rainy conditions lead to more frequent dashboard glances, longer fixation durations, and higher gaze elevation, indicating increased cognitive focus. These findings offer valuable insight into visual attention patterns under adverse conditions and highlight the potential of leveraging gaze modeling to aid design of more robust ADAS and DMS.
\end{abstract}

\keywords{Driver Gaze Behavior \and Adverse Weather Conditions \and Fixation Analysis \and Rainy Weather Driving \and Eye-Tracking \and ADAS \and DMS}

\section{Introduction}
The latest report from the World Health Organization~\cite{WHO} estimates that approximately 1.19 million people die each year due to road traffic accidents. These fatalities are attributed to a range of contributing factors, including distracted driving, driver fatigue, and inattention~\cite{huisingh2019distracted, tian2022driving}. Adverse weather conditions, particularly rain, also significantly increase to crash risk; for example, \cite{qiu2008effects, golob2003relationships} report that crash rates during rain can increase by up to 71\%. Specific conditions such as reduced visibility, interference from windshield wipers and raindrops, increased surface reflectivity, and diminished lane marking visibility contribute to the elevated risk. Collectively, these factors increase the cognitive demands of the dynamic driving task (DDT) and heighten the risk of accidents~\cite{huisingh2019distracted, tian2022driving}.

Given the automotive industry's current focus on developing feature-rich and robust ADAS technologies rather than making an immediate transition to Level 5 full automation~\cite{tengilimoglu2023implications}, it is critical to better understand human interactions with both the driving environment and emerging technologies. Although numerous studies have examined the behavior of various sensors within ADAS \cite{vargas2021overview, dreissig2023survey, farhani20243d}, far fewer have focused exclusively on human eye movements. Hence, one key element in this effort is the analysis of driver gaze behavior, which serves as a reliable proxy for attention and situational awareness~\cite{beauchemin2011portable}. Monitoring gaze behavior can yield valuable insights into driver readiness and cognitive engagement. Furthermore, understanding how experienced drivers adjust their gaze strategies under adverse conditions (e.g., rain) can inform the design and development of performant and robust human-machine interaction systems. In particular, expert gaze patterns can serve as a "teacher" model to improve the performance of ADAS.

Experienced drivers tend to concentrate their gaze on task-relevant and safety-critical regions of the visual field, primarily focusing on the far-road unless a salient object or hazard demands attention. In such cases, the gaze momentarily shifts before returning to the far-road view. This selective and efficient visual strategy supports safe driving by facilitating effective allocation of visual attention~\cite{lee2003computations, lappi2017systematic}.

While many studies have examined drivers’ gaze and facial cues \cite{rong2022and, jirjees2025integrated, tarba2020driver}, the question of how gaze strategies adapt to rainy conditions remains largely underexplored. Investigating variations in fixation duration, objects of interest, and visual search behavior under adverse weather can deepen our understanding of visual saliency and support the development of more accurate predictive models. These insights are crucial for advancing the design and effectiveness of ADAS.

In this work, we present a case study involving a single driver navigating under two distinct weather conditions: clear (sunny) and rainy. Eye gaze data were collected over two sessions with each lasting about 30-minute drive using a wearable gaze tracker (Neon Intelligent Glasses), which provides a data rate of 30 frames per second, resulting in approximately 54{,}000 annotated frames \((30~\text{minutes} \times 60~\text{seconds} \times 30~\text{fps})\). This dataset offers a rich stream of visual attention data.

The goal of this pilot study is to propose and validate a set of gaze-based metrics to capture meaningful behavioral differences between weather conditions. One key contribution of our work is the introduction of a clustering framework that segments gaze data into short time intervals, computes cluster centers, and aggregates them into meta-clusters—representing high-density gaze regions. We then construct a Markov transition matrix that captures the probabilistic transitions between these meta-clusters over time, providing a mathematical representation of gaze dynamics across weather conditions. Alongside this, we analyze other established metrics such as mean fixation duration, fixation duration distributions, and the azimuth/elevation distributions of gaze. Despite its single-subject scope, the high-resolution temporal data enable robust analysis and provide a framework for future large-scale studies.

The remainder of this paper is organized as follows. Section~\ref{sec:related_work} reviews prior work on eye gaze analysis under rainy conditions and summarizes commonly used gaze metrics while identifying gaps in the literature. Section~\ref{sec:methodology} details our data collection and methodological framework. Section~\ref{sec:Markov} introduces our novel clustering and Markov transition modeling approach. Section~\ref{sec:result} presents the experimental results, followed by a discussion in Section~\ref{sec:discussion} comparing gaze behavior across weather conditions. Finally, Section~\ref{sec:conclusion} concludes the paper by highlighting the significance and novelty of our approach and outlining directions for future research.

\section{Related Work}\label{sec:related_work}
In this section, we review previous research on driver gaze analysis under rainy conditions, followed by a discussion of the most commonly used comparison metrics in the literature. Finally, we highlight existing gaps in the field that the present study aims to address. 
\subsection{Driver's Gaze Analysis}
Studies examining driver gaze behavior under rainy conditions remain limited. One of the most relevant contributions is by \textit{Konstantopoulos et al.}~\cite{konstantopoulos2010driver}, who conducted a comprehensive driving simulator study to assess differences in gaze patterns between instructor-level (experienced) and learner drivers across various weather conditions, including daytime, nighttime, and rain. Their findings indicated that experienced drivers exhibited higher gaze sampling rates, shorter fixation durations, and broader scanning trajectories than learners. Nonetheless, the study also revealed that even experienced drivers exhibited a decline in visual search efficiency under low-visibility conditions, particularly during rainfall. The analysis was based on gaze-related metrics such as fixation count, mean fixation duration, and the standard deviation of horizontal and vertical fixation angles (in degrees).

In a related study focusing on foggy conditions rather than rain, \textit{Calsavara et al.}~\cite{calsavara2021effects} similarly found that visual search behavior became more constrained, with a narrowed field of view and an increased number of fixations. This study was also based on a simulated driving environment.

Building on these findings, \textit{Tavakoli et al.}~\cite{tavakoli2021environmental} conducted real-world driving experiments across clear, cloudy, and rainy conditions. They reported statistically significant changes in the standard deviation of gaze distribution in both horizontal and vertical directions between rainy and clear weather. However, their analysis was limited in scope, lacking further exploration of fixation behavior or comparative metrics.

Finally, \textit{Tian et al.}~\cite{tian2022driving} developed a deep learning model combining a convolutional neural network with a long short-term memory architecture to predict saliency maps under rainy conditions. While their method focused on modeling driver attention, it did not analyze how saliency or fixation patterns vary between different weather conditions. 

\subsection{Frequently Used Metrics of Comparison}

Several studies have utilized the number of fixations as a fundamental metric for comparison. The fixation is defined as the alignment of the the eyes on the fixed target for a given time (normally between 100 ms to 300 ms).  Mean fixation duration is another widely used measure, serving as a key indicator across various domains of eye gaze research, including fatigue analysis, attention and focus assessment, and weather-related studies~\cite{crundall2011visual, martens2007does}. In terms of scene analysis, the spatial distribution of gaze points overlaid on the driving environment has also been extensively examined~\cite{li2015driver, crundall2011visual, huang2021driver, konig2014nonparametric}. For instance, in a recent study, \textit{Huang et al.}~\cite{huang2021driver} employed semi-supervised K-means clustering to identify clusters of eye fixations on the scene during simulated driving scenarios. 

Gaze entropy, a metric grounded in Shannon’s information theory, quantifies the dispersion of visual attention across the scene, with higher entropy indicating more distributed fixations \cite{shiferaw2018stationary, di2016gaze}. This measure has been widely employed in various contexts, including the detection of alcohol-induced driver impairment and the classification of sleep-deprived individuals \cite{shiferaw2018stationary, shiferaw2019gaze}.

\subsection{Gap in Literature}

Despite the high frequency of traffic accidents occurring under rainy conditions, studies examining changes in driver gaze behavior in such weather are limited. Understanding how visual attention shifts in the rain can provide key insights into driver attention mechanisms. Of particular interest is identifying where drivers focus their attention, and how fixation points and patterns shift under adverse weather.

Previous studies have typically relied on a single metric, such as mean fixation duration, simple entropy measures, or clustering-based identification of areas of interest. However, no comprehensive study has been conducted to evaluate these metrics collectively—especially under rainy driving conditions. To address this gap, the present study aims to perform a multi-metric analysis of driver gaze behavior under both clear (sunny) and rainy weather. By comparing a range of gaze-related metrics, we seek to identify which features most effectively capture behavioral changes and scene analysis strategies. The results can inform the design of more robust and weather-aware ADAS models.

\section{Methodology and Material} \label{sec:methodology}
\subsection{Data Collection}

Data collection was carried out using Neon Intelligent Glasses, which record the driving scene at a frame rate of 30 frames per second, with each frame containing an overlaid point-of-gaze marker. The data were initially recorded using the companion mobile device provided with the glasses and were later transferred to a personal computer via the Neon Player Desktop application~\cite{Neon}. The driving route for this study spans approximately 24~km, extending from the NRC campus towards North via County Road~23, where the posted speed limit is 80~km/h. The driver participating in the study was experienced, and notably, the route corresponds to a regular commuting path taken on a weekly basis. As such, the driving environment was highly familiar to the driver, minimizing the influence of route unfamiliarity on gaze behavior. 

The rainy condition corresponds to a rain event on February 15, 2025, while the clear condition refers to data collected on June 27, 2025 that was a sunny day with no cloud cover. In this paper, the terms clear and sunny are used interchangeably, both referring to the data collected on June 27, 2025.


\subsection{DBSCAN Clustering} \label{sec:DBSCAN}

Density-Based Spatial Clustering of Applications with Noise (DBSCAN) is a widely used clustering algorithm that excels at identifying clusters of arbitrary shapes, including non-convex and irregular structures. Unlike K-means clustering, which requires the number of clusters to be specified in advance, DBSCAN does not rely on such prior assumptions. This makes it well-suited for applications such as identifying regions of interest in eye gaze data.

DBSCAN groups points based on local density using two key parameters:
\begin{itemize}
    \item $\epsilon$: the radius of the neighborhood around a point, and
    \item \textit{minPts}: the minimum number of points required within this neighborhood for a point to be considered a core point.
\end{itemize}

Points within high-density regions are grouped into clusters, while points in low-density areas that do not meet the criteria are labeled as noise. This density-based approach offers flexibility and robustness, making DBSCAN particularly suitable for gaze-based data analysis, where the number and shape of fixation clusters may vary naturally.

\subsection{Markov Transition Matrix} \label{sec:Markov}

A Markov process is a type of stochastic model in which the future state of a system depends solely on its current state, not on the sequence of events that preceded it. In this context, the system transitions between a finite set of states, and the probability of moving from one state to another is described by a transition probability matrix. Each entry $P_{i,j}$ in this matrix represents the probability of transitioning from state $j$ to state $i$.

In this study, we model the dynamics of eye gaze behavior as a Markov process, where the states correspond to distinct regions of visual attention—referred to as \textit{meta-clusters}. To construct these meta-clusters, we follow a multi-stage clustering approach:

\begin{enumerate}
    \item The continuous eye gaze data is first segmented into short, fixed-duration intervals (e.g., 10 seconds).
    
    \item Within each time chunk, DBSCAN is used to detect dense clusters of gaze points. The centroid of each cluster is calculated to represent the dominant area of visual focus for that interval. Notably, we do not explicitly differentiate between fixations and saccades. Instead, all gaze points are analyzed directly, under the assumption that regions of high spatial density inherently correspond to fixation-like behavior. This allows fixations to be inferred organically from the clustering process without manual classification.
    
    \item The resulting cluster centers from all time chunks are then pooled together and subjected to a second round of DBSCAN clustering to generate meta-clusters, which correspond to stable regions of visual attention over time.
    
    \item Each fixation point is identified in a time chunk is assigned to the nearest meta-cluster using a Euclidean distance criterion. As the gaze data evolves over time, this yields a temporal sequence of meta-cluster labels, effectively capturing the trajectory of gaze through distinct visual states.
    
    \item Finally, transitions between meta-clusters in successive time chunks are counted to build a transition matrix. This matrix is then normalized row-wise to obtain transition probabilities, reflecting the likelihood of the gaze shifting from one region of interest to another.
\end{enumerate}

\begin{figure}
    \centering
    \includegraphics[width=1\linewidth]{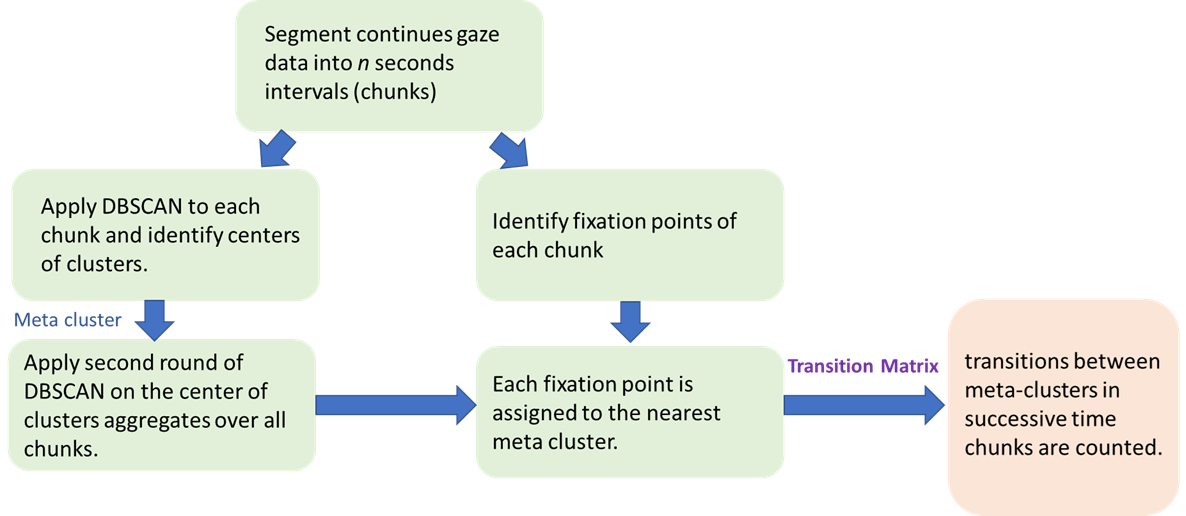}
    \caption{The Flow chart of the steps.}
    \label{fig:flowchart}
\end{figure}

The summary of mentioned steps as a flow chart is shown in \ref{fig:flowchart}. This approach enables us to quantify the temporal dynamics of gaze behavior and examine how transition patterns vary under different weather conditions. Modeling these transitions as a Markov process provides a framework to capture the driver's overall gaze strategy—specifically, the key regions of visual focus and the likelihood of shifting attention from one region to another. Given that gaze behavior has been shown in previous studies to exhibit repeatable and structured patterns, we anticipate that such regularities can be effectively captured and analyzed through the transition matrix representation.

\subsection{Fixation Duration}
\subsubsection{Fixation Duration Distribution}
The distribution of fixation durations can provide valuable insights into how visual attention differs between conditions (e.g., rain vs. clear weather). By analyzing the full distribution, rather than relying solely on summary statistics such as the mean and standard deviation, it is possible to capture more meaningful differences in gaze behavior. 

To statistically compare the distributions, the Kolmogorov–Smirnov (K-S) test can be employed. The K-S test evaluates whether two empirical distributions are likely to have originated from the same underlying population. Its advantage lies in being a non-parametric test, meaning it makes no assumptions about the form of the underlying distribution (in contrast to parametric alternatives such as the t-test).

\subsubsection{Fixation Duration Time Series}

Rather than computing a single aggregated mean fixation duration over the entire driving session, we propose segmenting the session into smaller time chunks (e.g., 10 seconds). For each segment, the mean fixation duration is calculated, resulting in a time series of mean fixation values. This approach not only facilitates intuitive visual comparison between conditions but also enables the computation of higher-level statistics, such as the mean and standard deviation of these segment-wise means.

\subsubsection{Coefficient of Variation}
The coefficient of variation (CV) is a statistical measure of relative dispersion that provides insight into the variability of a dataset. It is defined as $CV = \frac{\sigma}{\mu}$, where $\sigma$ represents the standard deviation and $\mu$ denotes the mean.
 
\subsubsection{Gaze Azimuth and Elevation Distribution}

In addition to analyzing fixation durations, the distributions of gaze azimuth and elevation provide valuable insights into how visual attention shifts under varying weather conditions. To evaluate whether these distributions differ significantly between clear and rainy scenarios, we employ the K–S test. To further quantify differences between the gaze distributions, we also use the Jensen–Shannon Distance (JSD), an information-theoretic metric that measures the similarity between two probability distributions. The JSD is symmetric and bounded between 0 and 1, where a value of 0 indicates identical distributions, and a value of 1 signifies completely disjoint distributions~\cite{lin2002divergence}. This dual approach—combining statistical hypothesis testing with information-theoretic divergence—enables a robust evaluation of gaze behavior differences across weather conditions.

\section{Result}\label{sec:result}
Based on the metrics defined in Section~\ref{sec:methodology}, we present the results of the weather-dependent gaze analysis. We begin with the clustering results for short time intervals, followed by the outcomes of meta-clustering and the construction of the Markov transition matrix. These are then complemented by the analysis of fixation duration distributions and other defined comparison metrics. 
\subsection{Clustering Result}
\begin{figure}
    \centering
    \includegraphics[width=1\linewidth]{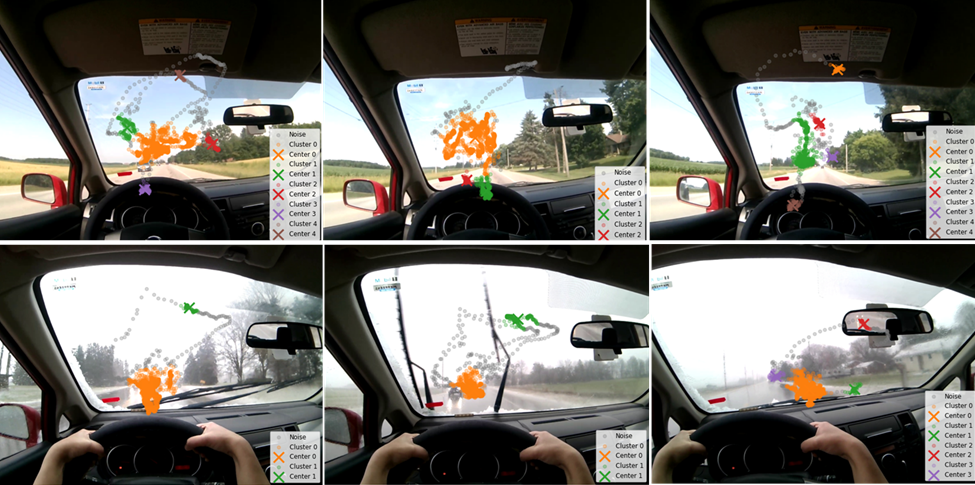}
    \caption{Three consecutive frames from the 10-second interval clustering analysis. \textbf{Top panel}: Clear weather condition. \textbf{Bottom panel}: Rainy weather condition.}
    \label{fig:single_cluster}
\end{figure}

The DBSCAN clustering applied to each 10-second segment of gaze data produced distinct clusters, with the coordinates of their centers extracted and stored for subsequent meta-cluster analysis, as outlined in Sec.~\ref{sec:Markov}. Fig.~\ref{fig:single_cluster} illustrates the clustering results, where each cluster is shown in a unique color, and the corresponding cluster centers are denoted by bold \textbf{X} markers. For visual reference, both the clustered gaze points and their respective centers are overlaid on the middle video frame, which corresponds to the 5-second mark within each 10-second interval. The top panel displays three consecutive 10-second segments recorded during a clear day, while the bottom panel presents the same sequence under rainy conditions.

Subsequently, all cluster centers from each 10-second segment were subjected to a second round of DBSCAN clustering to generate meta-clusters. These meta-clusters reflect the spatial distribution of the original cluster centers and provide an aggregated view of the driver’s major visual attention regions over the entire drive. 

Next, individual fixation points were identified using a duration threshold of 200~ms. Each fixation was then assigned to its nearest meta-cluster center based on Euclidean distance, as described in Sec.~\ref{sec:Markov}. This assignment enabled the construction of a Markov transition matrix that captures the temporal dynamics of gaze transitions between the dominant visual regions.

\begin{figure}
    \centering
    \includegraphics[width=0.6\linewidth]{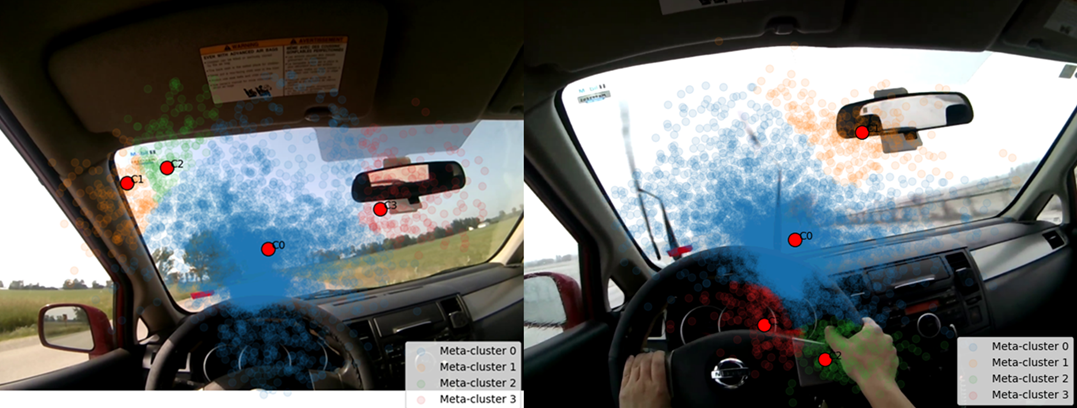}
    \caption{Meta-clustering results, where red circles denote the centers of individual clusters. \textbf{Left panel}: Clear weather. \textbf{Right panel}: Rainy weather.}
    \label{fig:meta_cluster}
\end{figure}

Fig.~\ref{fig:meta_cluster} shows the result of meta-clustering, with meta-cluster centers overlaid on a reference video frame for visualization purposes. While meta-clusters do not correspond to any specific frame, they offer a meaningful representation of the most frequently attended areas across all time segments.

In the clear (sunny) condition (left panel), four distinct meta-clusters are observed. The largest and densest cluster (blue) corresponds to sustained gaze on the road. Additional clusters appear in the rear-view mirror region (red) and the top-left area of the windshield (orange and green), indicating broader visual exploration.

Under rainy conditions (right panel), a similar large and dense cluster (blue) is centered on the road. However, the rear-view mirror region (orange) and the dashboard area (red and green) also exhibit a significant gaze concentration. This suggests that while the road remains the primary focus, adverse weather conditions may lead to increased attention toward vehicle instrumentation.

Furthermore, Fig.~\ref{fig:meta_cluster} presents the Markov transition matrices for both rainy and clear weather conditions. In both cases, the road remains the primary and most consistent focus of the driver’s gaze. Once the gaze is directed toward the road, it tends to remain there, with occasional transitions to other regions such as the rearview mirror, dashboard, or peripheral areas.

In clear weather, the driver's gaze is more exploratory. For instance, transitions to the upper-left regions of the windshield—labeled as UL~1 and UL~2 in the matrix—are more prominent. These correspond to the orange and green meta-clusters, respectively. In contrast, during rainy conditions, there is a noticeable shift in attention toward the dashboard area. The labels DB~1 and DB~2 in the matrix (associated with the red and green clusters) indicate this focus on dashboard-related regions.

Additionally, under clear conditions, there is evidence of gaze transitions within non-primary clusters—for example, between UL~1 and UL~2—suggesting a broader visual scanning behavior. This contrasts with the rainy condition, where gaze tends to remain more concentrated on road and dashboard areas, reflecting a more focused and possibly cautious gaze strategy.
\begin{figure}
    \centering
    \includegraphics[width=0.6\linewidth]{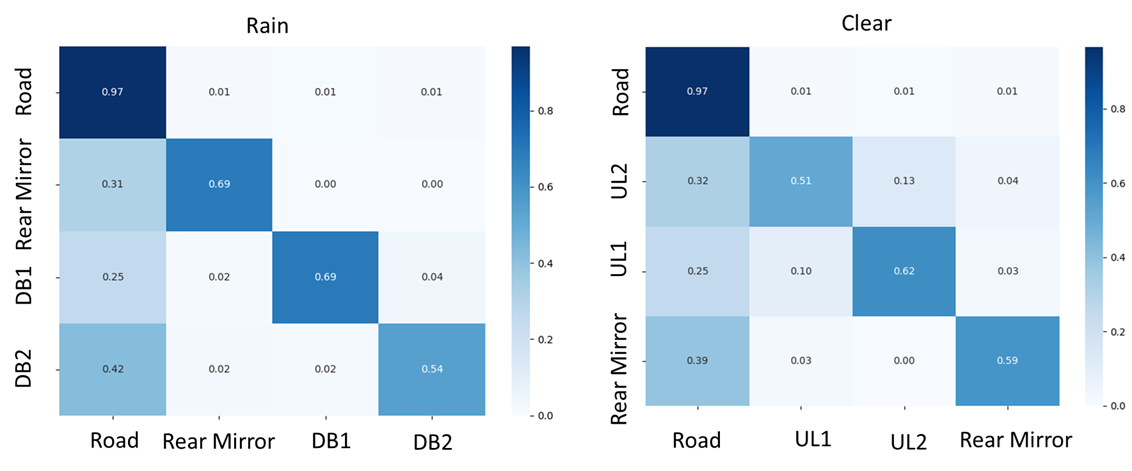}
    \caption{Markov transition matrices illustrating gaze dynamics under different weather conditions. \textbf{Left panel}: Rainy weather—labels DB~1 and DB~2 correspond to dashboard-related regions. \textbf{Right panel}: Clear weather—labels UL~1 and UL~2 correspond to the upper-left regions of the windshield.}

    \label{fig:markov_transition}
\end{figure}

\subsection{Fixation Duration Analysis}
\begin{figure}
    \centering
    \includegraphics[width=0.48\linewidth]{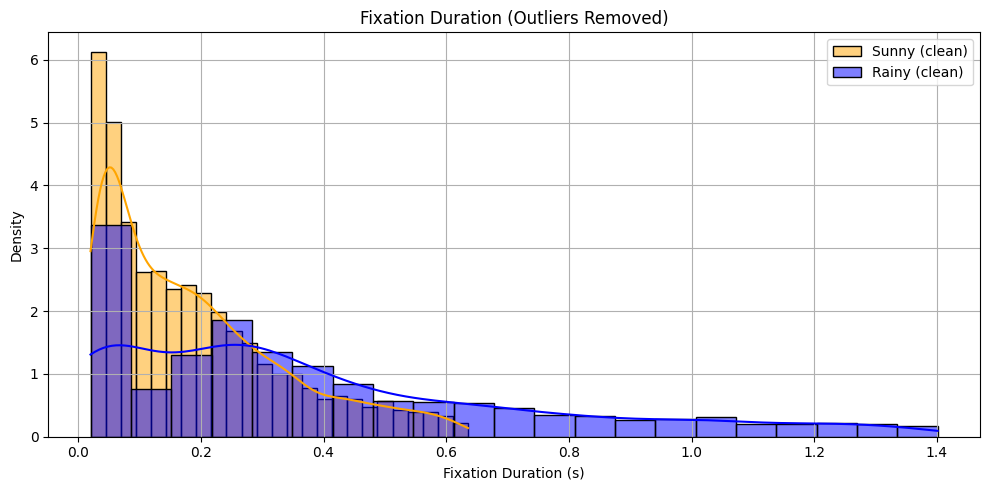}
    \includegraphics[width=0.48\linewidth]{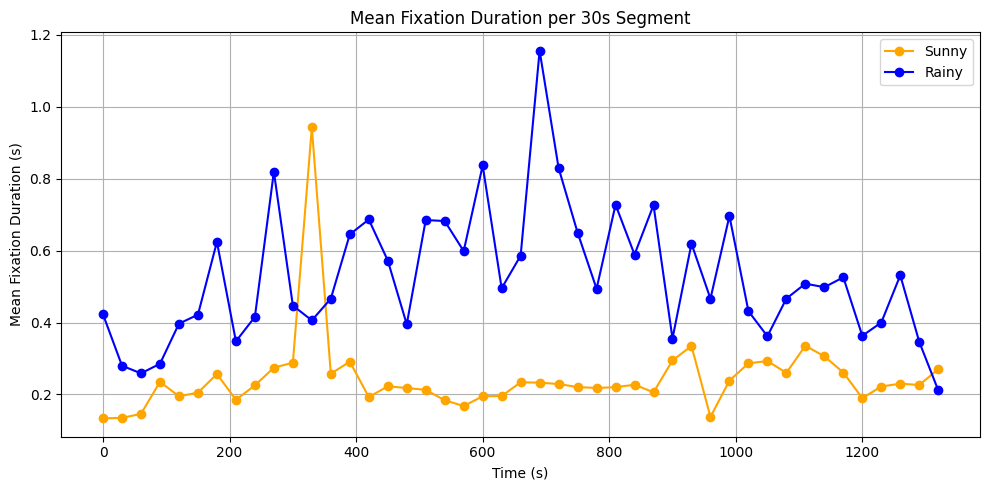}
    \caption{\textbf{Top panel}: Histogram of fixation durations. The orange curve represents clear (sunny) weather, while the blue curve corresponds to rainy weather. \textbf{Bottom panel}: Time series of mean fixation duration computed for each 30-second interval.}

    \label{fig:fixation_dist}
\end{figure}

The distribution of fixation durations for both rainy and sunny conditions is presented in the top panel of Fig.~\ref{fig:fixation_dist}. To avoid skewed results due to extreme outliers (i.e., unusually long fixation durations that may inflate the tails of the distribution), such values were removed prior to analysis. K–S test was conducted, yielding a $p$-value less than 0.0001, indicating that the two distributions are statistically distinct and likely originate from different underlying populations.

To enhance the visualization of temporal trends, the mean fixation duration was computed for each 30-second segment and plotted in the bottom panel of Fig.~\ref{fig:fixation_dist}. A longer interval was chosen to reduce the influence of outliers and ensure a more stable estimate of average fixation behavior. This segment-wise analysis highlights that fixation durations are, on average, significantly longer during rainy conditions. Specifically, under sunny conditions, the mean fixation duration is 0.25 seconds with a standard deviation of 0.12, whereas in rainy conditions, the mean increases to 0.53 seconds with a standard deviation of 0.19. CV further supports this observation: for sunny weather, CV = 0.48, while for rainy weather, CV = 0.35. This indicates that fixation durations are not only longer in rain but also more consistent, suggesting a shift toward more focused gaze behavior in adverse weather conditions.

\subsection{Gaze Azimuth and Elevation}
\begin{figure}
    \centering
    \includegraphics[width=0.5\linewidth]{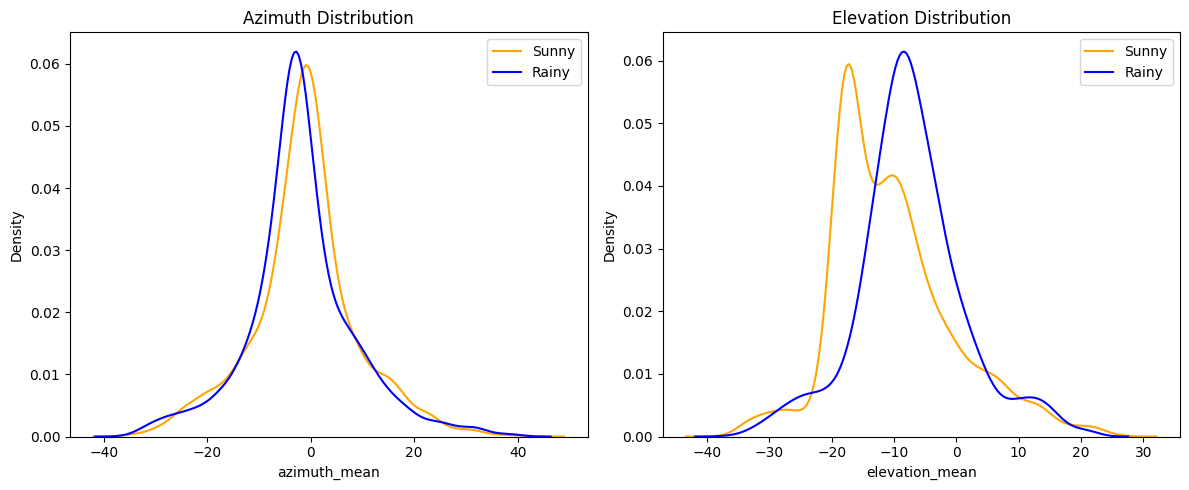}
    \caption{\textbf{Left panel}: Gaze azimuth distribution. \textbf{Right panel}: Gaze elevation distribution. The orange curve represents clear (sunny) weather, while the blue curve corresponds to rainy weather.}
    
    \label{fig:elevation}
\end{figure}

The distribution of gaze azimuth and elevation under both rainy and sunny conditions is illustrated in Fig.~\ref{fig:elevation}, where the orange curve represents sunny weather and the blue curve corresponds to rainy conditions. A Kolmogorov–Smirnov (K–S) test performed on both azimuth and elevation data revealed statistically significant differences between the two weather conditions ($p < 0.0001$).

Although the azimuth distributions for sunny and rainy conditions are relatively similar and approximately Gaussian—centered at $-1.06^\circ$ and $-1.85^\circ$, respectively—the elevation distributions exhibit more notable divergence. Under sunny conditions, a clear bimodal pattern is observed, while the rainy condition shows a single, higher elevation focus, with a mean of $-7.34^\circ$ compared to $-10.26^\circ$ in clear weather. The Jensen–Shannon Distance (JSD) between the two-dimensional gaze distributions (azimuth and elevation) is 0.25, indicating a moderate shift in gaze behavior across weather conditions.

\section{Discussion}\label{sec:discussion}
The results presented in Section~\ref{sec:result} reveal several noteworthy insights. The clustering analysis, along with the corresponding Markov transition matrices, indicates that gaze behavior exhibits a high degree of repeatability across both clear and rainy weather conditions—consistent with prior qualitative findings based on observational studies. In particular, the driver’s primary focus remains directed toward the road ahead, with occasional gaze shifts toward the rear-view mirror to monitor traffic from behind. While individual clusters derived from 10-second intervals (see Fig.~\ref{fig:single_cluster}) exhibit some scene-dependent variability, the road consistently appears as one of the primary gaze targets.

It is important to note that the clustering over short time intervals is performed on all gaze points without explicitly distinguishing between fixations and saccades. This approach is based on the assumption that high-density regions inherently correspond to areas where the eyes fixate for longer durations. However, once the meta-clustering is applied—aggregating the centers of these short-interval clusters over time—only the fixation points are retained and assigned to the resulting meta-clusters. This refinement allows for a more targeted analysis, as the focus is specifically on fixation behavior.

The meta-clustering analysis, which captures long-term patterns in gaze behavior, provides additional insight. Notably, under rainy conditions, the driver’s gaze shifts more frequently toward the dashboard, likely to monitor essential vehicle information such as speed, fuel level, engine RPM, or to interact with controls related to windshield wipers and defogging. In contrast, during clear weather, gaze more frequently shifts toward the upper-left region of the visual field—potentially reflecting engagement with peripheral environmental stimuli such as clouds or birds—indicating a more relaxed and exploratory visual scanning strategy.

The Markov transition matrices further support these observations: in both weather conditions, the dominant transition probability remains associated with maintaining gaze on the road, indicating that the fundamental visual strategy of forward attention is preserved. However, rainy conditions introduce a secondary visual strategy characterized by increased attention to dashboard elements, reflecting a compensatory response to the added demands of inclement weather.

Fixation duration analysis reveals a substantial difference between conditions: on average, fixations in rainy weather are approximately twice as long as those in clear weather, indicating increased cognitive load and attentional focus. Complementing this, the CV for fixation durations is higher under clear conditions, suggesting a more variable and exploratory visual behavior.

Additionally, gaze elevation is found to be higher in rainy conditions, implying that the driver is looking farther down the road—potentially as a compensatory mechanism for reduced visibility. This adaptation may enhance the ability to anticipate upcoming obstacles or changes in road conditions. Finally, the two-dimensional JSD between the azimuth-elevation distributions across weather conditions is 0.25, reflecting a moderate shift in overall gaze behavior.

Together, these findings indicate that while the fundamental driving gaze strategy remains stable, weather-related environmental constraints elicit specific adaptive behaviors in gaze distribution, duration, and elevation.

\section{Conclusion and Future Direction} \label{sec:conclusion}
In the present work, we presented the results of a case study investigating alterations in gaze behavior under rainy weather conditions compared to clear weather. We introduced a novel two-step clustering framework for time series gaze data, where the first step segments the data into 10-second intervals and identifies high-density gaze clusters. The centroids of these clusters are then used to construct meta-clusters that represent the dominant gaze targets over time. We demonstrated that this clustering approach, when combined with Markov transition matrices, offers a mathematical representation of gaze strategies that aligns with previous qualitative studies, which suggest that experienced drivers maintain a highly repeatable gaze pattern, with the majority of attention focused on the road~\cite{lappi2017systematic}.

The proposed clustering framework offers the advantage of capturing both short-term gaze dynamics—indicating where the driver is looking in each moment—and long-term trends that reveal overarching gaze strategies. A natural extension of this pilot study involves more detailed analysis of short-term clusters across varying driving scenarios (e.g., intersections, lane changes, and roundabouts), which may uncover context-specific gaze adaptations under different weather conditions.

In addition to clustering, we examined several complementary metrics, including fixation duration distributions, time series of mean fixation durations within each interval, and the joint azimuth–elevation distributions. These metrics served as strong indicators of differences in gaze behavior between weather conditions. 

This case study revealed several key findings. While the general gaze strategy remained consistent across weather conditions—with drivers primarily focused on the road—in rainy weather, gaze shifts toward the dashboard occurred more frequently, likely reflecting increased monitoring of vehicle status. In contrast, during clear weather, the driver’s gaze exhibited more exploratory behavior. Notably, the mean fixation duration in rainy conditions was nearly twice as long as in clear weather, indicating increased cognitive load and attentional demand.

This work opens multiple avenues for future research. Expanding the dataset to include multiple drivers and extended recording sessions could help assess inter-individual variability in gaze strategies. Additionally, while the current study focused solely on gaze data, incorporating pupil dilation could provide further insight into driver alertness and cognitive state. Head movement data may also reveal behavioral variability not captured by gaze alone. Exploring appropriate metrics to quantify and combine these multimodal signals represents a promising direction for enhancing our understanding of driver attention and readiness in complex, real-world driving environments.

\bibliographystyle{unsrtnat}
\bibliography{references}  






\end{document}